\documentclass[twoside, 11pt]{article}
\usepackage{jair, theapa, rawfonts}
\usepackage{amsmath}
\usepackage{amsfonts}
\usepackage{dsfont}
\usepackage{algorithm}
\usepackage{algorithmic} 
\usepackage{graphicx}
\usepackage{chngcntr}
\usepackage{url}
\usepackage{cases}

\counterwithout{figure}{section}
\newtheorem{conjecture}{Conjecture}

\ShortHeadings{Planted Dense Subgraphs Can Be Recovered using Graph-based Machine Learning}{Levinas \& Louzoun}
\jairheading{75}{2022}{541-568}{10/2022}{06/2022}
\title{Planted Dense Subgraphs in Dense Random Graphs Can Be Recovered using Graph-based Machine Learning}

\author{\name Itay Levinas \email levinai@biu.ac.il \\ \addr
    Department of Mathematics, Bar Ilan University\\ Ramat Gan, Israel
    \AND
    \name Yoram Louzoun \email louzouy@math.biu.ac.il \\ \addr Department of Mathematics, Bar Ilan University\\ Ramat Gan, Israel
}

\begin{document}
\maketitle

\begin{abstract}
Multiple methods of finding the vertices belonging to a planted dense subgraph in a random dense $G(n, p)$ graph have been proposed, with an emphasis on planted cliques. Such methods can identify the planted subgraph in polynomial time, but are all limited to several subgraph structures. Here, we present PYGON, a graph neural network-based algorithm, which is insensitive to the structure of the planted subgraph. This is the first algorithm that uses learning tools for recovering dense subgraphs. We show that PYGON can recover cliques of sizes $\Theta\left(\sqrt{n}\right)$, where $n$ is the size of the background graph, comparable with the state of the art. We also show that the same algorithm can recover multiple other planted subgraphs of size $\Theta\left(\sqrt{n}\right)$, in both directed and undirected graphs.
We suggest a conjecture that no polynomial time PAC-learning algorithm can detect planted dense subgraphs with size smaller than $O\left(\sqrt{n}\right)$, even if in principle one could find dense subgraphs of logarithmic size. 
\end{abstract}

\section{Introduction}
Let $G = (V, E)$ be a graph, where $V$ is a set of vertices and $E \subseteq V\times V$ is a set of edges. A clique in $G$ is a subset of $V$, where each vertex is connected to all the other vertices in the subset. The problems of finding, or even determining, the size of the maximum clique in a graph, are fundamental problems in theoretical computer science and graph theory, and are known to be NP-hard \cite{Karp72}. Let $G(n, p)$ denote the collection of random (undirected) Erd\H{o}s-R\'enyi graphs, each of which is a graph with $n$ vertices, such that each edge $\{i, j\}$ exists in the graph with probability $p$, independent on all other edges\footnote{We also consider the case of directed graphs, with the same notation, in which each edge is an ordered pair $(i, j)$ and exists with probability $p$ independent on all other edges.}. Asymptotically almost surely (i.e. with probability that approaches 1 as the size of the graph, denoted by $n$, tends to infinity), the size of the maximum clique in a $G(n, p)$ graph is roughly $2 \log_{\frac{1}{p}}{n}$, however no polynomial time algorithm is known to find a clique of size $(1 + \Theta\left(1\right)) \log_{\frac{1}{p}}{n}$. The problem of finding a clique of size $(1 + \epsilon)  \log_{2}{n}$ in a $G(n, \frac{1}{2})$ for any $\epsilon>0$ in polynomial time was suggested by Karp \citeyear{Karp76}.

We focus on a tad easier problem: the \textit{planted}, or \textit{hidden clique problem}, suggested independently by Jerrum \citeyear{Jerrum92} and Ku\v{c}era \citeyear{Kucera95}. Let $G(n, p, k)$ denote the collection of $G(n, p)$ graphs, where we select random $k$ vertices and replace the subgraph induced by them with a clique. The problem is to find a polynomial time algorithm that, given a $G(n, p, k)$ graph (where $k$ is known), recovers the clique. The common case in the literature is $G(n, \frac{1}{2}, k)$ and as such, most algorithms were addressed to edge probability in particular, but could easily be extended to more general $G(n, p, k)$ for some constant $p$.

The planted clique and related problems have important applications in a variety of areas, including community detection \cite{HWX15},  molecular biology \cite{PS00},  motif discovery in biological networks \cite{MSO+02,JM15}, computing the Nash equilibrium \cite{HK11,ABC13}, property testing \cite{AA+07}, sparse principal component analysis \cite{BR13}, compressed sensing \cite{KZ14},
cryptography \cite{JP00,ABW10}, and even mathematical finance \cite{AB+11}. 

A simple quasi-polynomial time algorithm that recovers maximum cliques of any size $k \geq 2 \log_{\frac{1}{p}}{n}$ is to enumerate subsets of size $2 \log_{\frac{1}{p}}{n}$; for each subset that forms a clique, take all common neighbors of the subset; one of these will be the planted clique. This is also the fastest known algorithm for any $k = O(n^{\frac{1}{2} - \delta})$, where $\delta > 0$.

There are also polynomial time algorithms, but up to now they are able to recover only cliques of size at least $\Theta\left(\sqrt{n}\right)$. Ku\v{c}era \citeyear{Kucera95} observed that when $k > c\sqrt{n \log\left(n\right)}$ for an appropriate constant $c$, the vertices of the planted clique would almost surely be the ones with the largest degrees in the graph, therefore this case is easy. The first polynomial algorithm that finds, asymptotically almost surely, cliques of size $c\sqrt{n}$, where $c$ is sufficiently large, was presented by Alon, Krivelevich and Sudakov \citeyear{AKS98}\footnote{They also suggested a polynomial time algorithm to find planted cliques of size $c\sqrt{n}$ for any $c > 0$, which are polynomial time algorithms with degree that grows as $c$ decreases. Our comparison of $c$ values here is between algorithms with computation time smaller than $O(n^3)$.}, using spectral techniques. Many other algorithms further improved $c$. In parallel, lower bounds for clique sizes able to be found using some families of algorithms, have been proven over the years. An important lower bound, first reached in an algorithm of Deshpande and Montanari \citeyear{DM15}, is $k=\sqrt{\frac{n}{e}}$. To the best of our knowledge, this is the lowest bound to which state of the art algorithms compare.

In the regime $\frac{k}{\sqrt{n}}\to{0}$, various computational barriers have been established for the success of certain classes of polynomial-time algorithms, such as the Sum of Squares Hierarchy \cite{BH+19}, the Metropolis Process \cite{Jerrum92} and statistical query algorithms \cite{FG+17}. Gamarnik and Zadik \citeyear{GZ19} have presented evidence of a phase transition for the presence of the Overlap Gap Property (OGP), a concept known to suggest algorithmic hardness when it appears, at $k = \Theta (\sqrt{n})$. Moreover, the assumption that detecting a planted clique of size $k = o(\sqrt{n})$ in polynomial time is impossible has become a key assumption in obtaining computational lower bounds for many other problems \cite{HWX15,BR13,BBH18,BB18}. However, no general algorithmic barrier such as NP-hardness has been proven for recovering the clique when $k = o(\sqrt{n})$.

Here, we show a novel graph convolution network-based algorithm to recover cliques of sizes $\Theta\left(\sqrt{n}\right)$, with similar performance as Deshpande and Montanari \citeyear{DM15}. The algorithm has two stages - a learning stage that outputs vertices highly likely to belong to the planted clique, and a cleaning stage that helps obtaining the planted clique based on the candidate vertices. To the best of our knowledge, there is no advanced machine learning tool to detect planted cliques. We show that the algorithm performs as well for recovering planted dense subgraphs other than cliques (e.g. $K$-plexes and $G(k, q)$ for $q > p$) and for directed graphs. In addition, although focusing mostly on $G(n, \frac{1}{2})$ (as most common papers do), we show that our algorithm works also if $p$ is a constant other than $\frac{1}{2}$. We thus produce state of the art accuracy for planted clique recovery and enlarge the planted clique problem to the dense subgraph recovery problems.    

\section{Related Work} \label{related work}
The planted clique problem has been studied extensively. We discuss here algorithms relevant to our work.

As mentioned above, the first polynomial time algorithm was introduced by Alon et al. \citeyear{AKS98}, for cliques of sizes $c \sqrt{n}$ where $c$ is large enough. In a variant of this algorithm\footnote{This variation is used by Deshpande and Montanari \citeyear{DM15}. The original variation uses the plain adjacency matrix and its second eigenvector.}, one can use the following representation of the adjacency matrix:
\begin{subnumcases}{A_{ij} = \frac{1}{\sqrt{n}}}
1, & $i\neq j \text{ and} \left\{i, j\right\} \in E$ \\
0, & $i = j$ \\
-f(p), & $i\neq j \text{ and} \left\{i, j\right\} \notin E$
\end{subnumcases}
where $f(p)$ is a positive function of $p$ (where $f(0.5)=1$, to adjust this algorithm, originally proposed for $G(n, \frac{1}{2})$, to any $G(n, p)$).
If $c$ is large enough, then the clique can be recovered using the entries of the eigenvector corresponding to the largest eigenvalue of $A$. Selecting the $k$ largest entries (by absolute value) of this eigenvector yields a subset of vertices, that includes at least half the vertices of the planted clique; the other half can be subsequently identified through the following simple procedure -- choosing the vertices connected to at least $\frac{3k}{4}$ vertices in the previous subset.

Ron and Feige \citeyear{FR10} used a method of iterative removal of vertices, based on values related to the degrees of each vertex. With $O(n^2)$ complexity, it succeeds with probability $\frac{2}{3}$ when $k \geq c \sqrt{n}$, for sufficiently large (unspecified) $c$. 

Ames and Vavasis \citeyear{AV11} studied a convex optimization formulation in which the graph's adjacency matrix was approximated by a sparse low-rank matrix. These two objectives (sparsity and rank) are convexified through the usual $l_1$-norm and nuclear-norm (i.e., sum of the singular values of the matrix) relaxations. They proved that this convex relaxation approach is successful asymptotically almost surely, provided $k \geq c \sqrt{n}$ for a
constant $c$ unspecified in the paper.

Other works were based on approximating the Lov\'asz theta function \cite{Lovasz79}. An algorithm proposed by Feige and Krauthgamer \citeyear{FK00}, is able to find the clique asymptotically almost surely for $k \geq c \sqrt{n}$ if $c$ is large enough, by maximizing a function of orthogonal vectors corresponding to each vertex, which is an approximation of the Lov\'asz theta function of the graph. Jethava et al. \citeyear{JMBD13} then solved analytically the Support Vector Machine (SVM) problem to approximate  the Lov\'asz theta function. Their algorithm works for graphs with edge probability $\frac{\log^4{n}}{n}\leq p < 1$ and clique sizes $k \geq 2 \sqrt{\frac{1-p}{p} n}$.

Dekel et al. \citeyear{DGP14} improved the algorithm of Feige and Krauthgamer \citeyear{FR10}, to recover cliques of sizes $k \geq 1.261 \sqrt{n}$ with time complexity of $O(n^2)$. The improved algorithm includes three phases: First, create a sequence of subgraphs of the input graphs $G$, $G=G_0 \supset{G_1} \supset ... \supset{G_t}$, where $G_{i+1}$ is created from $G_i$ based on the degrees of vertices in $G_i$. Then, find the subset of the clique contained in $G_t$, denoted by $\Tilde{K}$. In the last phase, looking at the subgraph of $G$ induced by the vertices of $\Tilde{K}$ and their neighbors, the candidate clique vertices are the $k$ vertices there with the highest degrees.

Deshpande and Montanari \citeyear{DM15} used a belief propagation algorithm, also known as approximate message-passing. This algorithm is, to the best of our knowledge, the best proven polynomial time algorithm, recovering cliques as small as $\sqrt{\frac{n}{e}}$ in theory, with time complexity of $O(n^2\log{n})$. The main idea of this algorithm is passing "messages" between vertices and their neighbors, normalizing and repeating until the messages on each vertex converge to the probability that the vertex belongs to the clique. In more detail, this message passing algorithm includes two main phases. The first is the message passing phase described above, after which one receives the likelihood for each vertex to belong to the clique. One builds a set of candidate vertices that are the most likely to belong to the clique, but this set, although containing many clique vertices, is much larger than $k$. Therefore, a second phase of cleaning, similar to the algorithm shown by Alon et al. \citeyear{AKS98}, is applied to recover the clique. The concept of belief propagation was used in other tasks beyond the hidden clique recovery, such as dense subgraph recovery \cite{HWX18} graph similarity and subgraph matching \cite{KP+11}.

The algorithm of Deshpande and Montanari was extended through statistical physics tools by Chiara \citeyear{Angelini18}. First, Chiara computed the Bethe free energy associated to the solution of the belief propagation algorithm. Then, the range of clique sizes was separated into intervals where the algorithm can or cannot be used for solving the problem, by analyzing the minima of the free energy function. Lastly, an algorithm of Parallel Tempering was applied to find the cliques where belief propagation struggles to do so. Although the Parallel Tempering algorithm recovers cliques smaller than $\sqrt{\frac{n}{e}}$, the algorithm does so with higher time complexity\footnote{For instance, according to the values found in the paper, the time complexity when $k = \sqrt{n}$ is proportional to $n^{3.96} \log_2^{3.64}{n}$, longer by orders of magnitude than existing algorithms (including the algorithm we propose).}, that diverges (even in the average case) as the clique size approaches the threshold size of cliques in $G(n, \frac{1}{2})$ graphs. 

The algorithm proposed by Marino and Kirkpatrick \citeyear{MK18} further improved a greedy local search algorithm. Using search algorithms of at least $O(n^3)$ computation time, a greedy search for a clique of size larger than the size of a typical maximum clique in a $G\left(n, \frac{1}{2}\right)$ was performed. Then, there was applied a cleanup operation (which is used as an early stopping mechanism) to recover the full clique from the smaller one. Marino and Kirkpatrick show that they can recover cliques smaller than $k=\sqrt{\frac{n}{e}}$ in polynomial time. However, they did not prove that the early stopping mechanism reduces the complexity of their algorithms, so even with the early stopping, these algorithms may still take $O(n^3)$ time or more.

Machine learning on graphs has been used to tackle computationally hard graph problems, for instance by Khalil et al. \citeyear{KDZ+17}. They used a combination of graph embedding with reinforcement learning to give approximated solutions for the Minimum Vertex Cover, Maximum Cut and Traveling Salesman problems. However, our algorithm solves a the slightly different problem of recovering planted subgraphs, being the first to do so using such tools and providing an exact solution when successful. 

\section{Methods}
This section includes a detailed explanation of our algorithm, including the approach we take to recover the planted subgraphs.

\subsection{Algorithm Outline}
We follow the methods above and use a two stage approach. At the first stage, we detect a subset of vertices belonging, with a high probability, to the planted subgraph. At the second stage, we extend and clean this subset to obtain the full subgraph. We here mainly focus on the first stage, since the second stage is similar to previous methods.

For the first stage, we propose a framework based on machine learning on graphs, which we address by the name PYGON (\textbf{P}lanted Subgraph recover\textbf{Y} using \textbf{G}raph c\textbf{O}nvolutional \textbf{N}etwork)\footnote{Our implementation of PYGON may be found at:  \url{https://github.com/louzounlab/PYGON}.}. 
In short, we produce random realizations of the required planted subgraph recovery task, and train a model based on Graph Convolutional Networks (GCN) \cite{KW16} to find the planted subgraph, using topological features of the vertices as an input to the GCN. We then compute the prediction accuracy on new realizations of the planted subgraph task in different random graphs. 
The advantage of PYGON is that a change in the target subgraph type only leads to a change in the simulated subgraph planting. The weights of the predictor for the presence of such  subgraphs are then learned through a machine learning formalism. In the following methods sections, we describe the details of the machine learning algorithm and the input used for each vertex.

\subsection{Model Structure}
Here, we describe the building blocks that we combine to create PYGON. As we describe below, some building blocks are replaceable with similar blocks. We tested several such replacements in the Results Section and Appendices \ref{Appendix B} and \ref{Appendix C}.

\subsubsection{Setup}
Given the desired graph size $n$, edge probability $p$ and planted subgraph size $k$, we train PYGON using known examples. We generate $G(n, p)$ graphs, for each of which we select a random set of $k$ vertices, and set the subgraph induced by them as the pattern we desire\footnote{When the subgraph may have more than one possible pattern, e.g. a $G(k, q)$ subgraph, we randomly select a pattern for each graph and plant this pattern in the selected vertices.}. Note that the training, evaluation and test graphs have the same size, edge probability and planted subgraph size. The graphs are separated into training and evaluation sets. We label the vertices, such that the planted subgraph vertices are labelled 1 and vertices not in the planted subgraph are labelled 0. We then build a model to predict the label of the vertices in all input graphs. 

\subsubsection{Features per Vertex}\label{Features per Vertex}
We use an input matrix of vertex features for each graph, $X\in \mathbb{R}^{n\times f}$, such that $X_{i,j}$ is the value of the $j$-th feature of the $i$-th vertex. The features we tested are of relatively low complexity, describe local properties of each vertex and known to be effective in graph topology-based machine learning tasks \cite{ABL19,NCL19}. Note that PYGON must receive a feature matrix representing each input graph, including test graphs, but can be used even with no computed features, and instead use an identity matrix as $X$. However, better performance was obtained using calculated features. The features are calculated using a package we developed\footnote{https://github.com/louzounlab/graph-measures}.

The features we tried are the following:
\begin{itemize}
    \item{\textbf{Degree - }}We used this feature in both directed and undirected cases, where in the directed case we considered the degree as the total degree (i.e. number of ancestors and descendants). However, one may use in- or out-degrees, or any combination of them.
   
    \item{\textbf{3-Motifs - }}The $i$-th 3-motif value for a vertex $v$ is the number of different triplets of vertices to which $v$ belongs and that the subgraph induced by them is isomorphic to the $i$-th type of 3-motif. Using a novel method for motif enumeration \cite{LSL22}, implemented in our package, we can calculate the motifs per vertex with time complexity of $O(n^3)$, but distributing the computation on a Graphics Processing Unit (GPU) significantly shortens the computation time.
\end{itemize}

\subsubsection{Modified Adjacency Matrix}
From the original adjacency matrix of each graph, we develop a modified matrix with learnable elements, which is used in the learning process - $\tilde{A}\in\mathbb{R}^{n\times n}$, such that:
\begin{subnumcases}{\label{eq:adjacency matrix}\tilde{A}_{ij} = \frac{1}{\sqrt{n}}}
    \gamma & $i = j$ \\
    \frac{1-p}{p} e^{\alpha} & $i \neq j \text{ and } (i, j) \in E$ \\
    -e^{\beta} & $i \neq j \text{ and } (i, j) \notin E$
\end{subnumcases} 
Where $\alpha, \beta, \gamma$ are coefficients, affecting the weight of information passing through a vertex to its neighbors, to non-adjacent vertices and to be preserved at the vertex itself, respectively. We let $\beta$ and $\gamma$ be learned during the training process and keep $\alpha$ constant. We initialize the coefficients so that $e^{\alpha} = 1$, $e^{\beta} = 1$ and $\gamma = -1$. \\
We multiply $e^\alpha$ by $\frac{1-p}{p}$ because when taking $p \neq \frac{1}{2}$, the expected number of adjacent vertices is different from the number of disjoint vertices, creating a severe bias in the model (See \ref{3.2} for more information regarding the logic behind the modifications in the adjacency matrix).

\subsubsection{Graph Convolutional Network} PYGON is based on a multiple-layer Graph Convolutional Network (GCN). A GCN layer \cite{KW16} is a graph neural network layer that learns feature representations of vertices, based on their input features and the topology of the graph. Formally, if $H^{(l)} \in \mathbb{R}^{n\times f_{l}}$ represents the feature matrix input to the $(l+1)$-th layer and the graph is represented by a matrix $\tilde{A}$ (usually a modified version of the adjacency matrix), then the GCN layer outputs a new feature representation for the vertices:
\begin{equation}
    H^{(l+1)} = \sigma\left(\tilde{A}\cdot H^{(l)} \cdot W^{(l+1)}\right)
\end{equation}
where $\sigma$ is a nonlinear activation function and $W^{(l+1)} \in \mathbb{R}^{f_{l}\times f_{l+1}}$ is the learnable weight matrix of the $(l+1)$-th layer.

\subsubsection{Learning Model}
\begin{figure*}[!t]
    \centering
    \includegraphics[width=15cm]{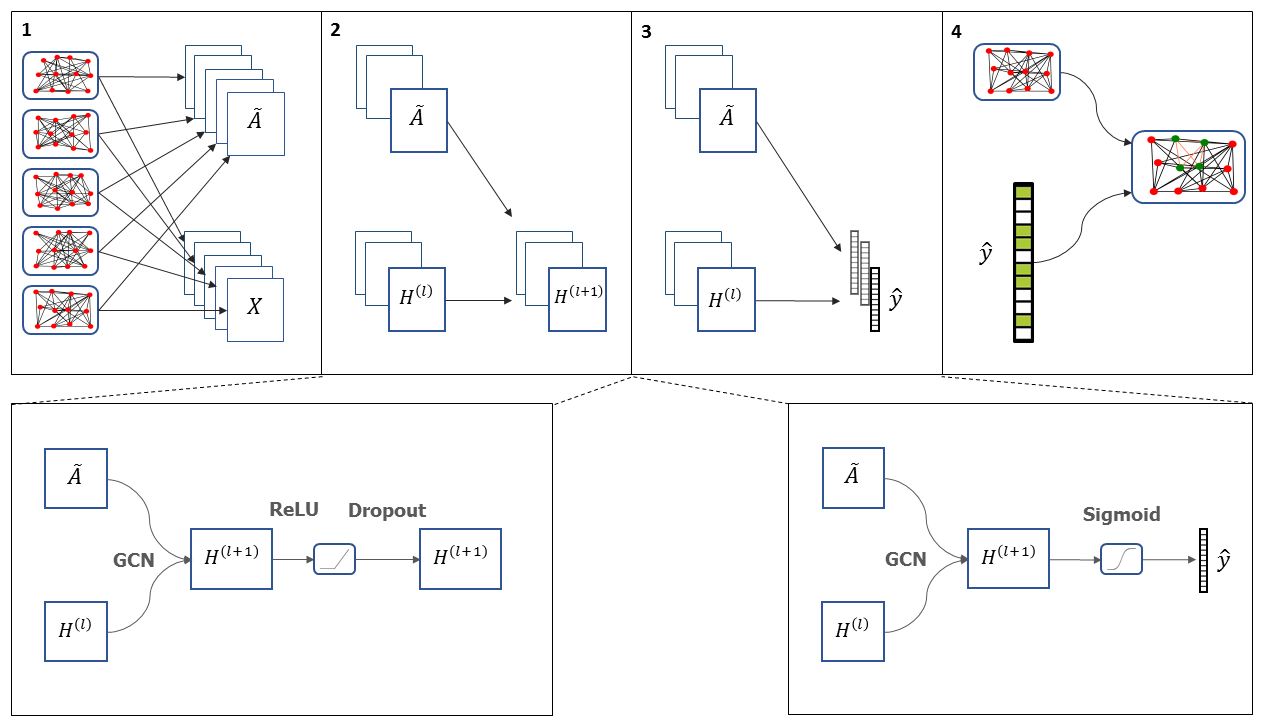}
    \caption{
        The end-to-end structure of our model, made of four parts: 1. Feature calculations -- calculating a feature vector for each vertex to serve as an input to the learning model. 2. Hidden GCN layers -- Feature and adjacency matrices are input to the GCN layers, obtaining a new feature representation of the vertices. Then, this representation is passed through a $ReLU$ activation function and a dropout layer. The lower left frame shows the process of one hidden layer. 3. Final layer -- Feature and adjacency matrices are input to the final GCN layer, producing a column vector. This vector is passed through a softmax function to obtain a vector associated with the predicted probabilities for each vertex to belong to the planted subgraph. The lower right frame shows the process of the final layer. 4. Cleaning procedure -- a general name for the stage from selecting the vertices considered as candidates to belong to the subgraph, by the outputs of PYGON, to fully recovering the subgraph.
        }
    \label{fig:model structure}
\end{figure*}

PYGON receives as input an initial feature matrix $X$ and an adjacency-like matrix $\Tilde{A}$, and outputs a prediction vector $\hat{y}\in\left[0, 1\right]^n$. 
The inputs are fed into a GCN layer with a $ReLU(x) = \max{(0, x)}$ activation function and then a dropout layer, to receive the input features to the next GCN layer. After the desired number of such layers, the last layer is a GCN with a sigmoid function (without dropout), that outputs $\hat{y}$. The architecture of the model (PYGON and the extension stage) is presented in Figure \ref{fig:model structure}.

To train the model to predict vertices belonging to the desired planted subgraph, we aim to minimize a weighted binary cross-entropy loss function (later denoted as $WBCE\left(\hat{y},\ y\right)$:
\begin{equation}\label{eq:model loss}
    L\left(\hat{y},\ y\right) = -\sum_{i=1}^{N} {\frac{1}{k} y \log{\hat{y}} + \frac{1}{n-k} \left(1 - y\right) \log{\left(1 - \hat{y}\right)}}
\end{equation}
Different loss functions have been tried, as shown in detail in appendix B, yet this function has shown the best performance. 

\subsection{Learning Process}
We here describe some technicalities in our learning process: our training in epochs, early stopping mechanism and hyper-parameter selection. 

\subsubsection{Epochs}
PYGON is trained in epochs. At every epoch, we shuffle the order of graphs and train the model by them, one graph at a time. In other words, the backpropagation training process is done multiple times per epoch, once for each training graph. Our motivation in doing so is similar to the motivation for batched stochastic gradient descent is used in many machine learning tasks: On one hand, each graph alone is a unit made of many vertices, making the training step from it quite robust, and on the other hand, we wanted to make as many "random" training steps as we can to converge fast.

\subsubsection{Early Stopping}
We set a maximal number of epochs to train PYGON, but it can be stopped earlier. During the training process, we evaluated the performance on the evaluation set, and if for 40 epochs the loss on the evaluation set has not decreased below the current minimal loss, the training was stopped. Finally, our trained model is the model in the epoch that gave the best evaluation loss. 

\subsection{Hyper-parameter Selection}
We tuned the hyper-parameters of PYGON using  Neural Network Intelligence (NNI) \cite{NNI}. NNI helps choosing the best model hyper-parameters out of a specified range of values, based on many trials of training the model. Each performance trial is done using a different hyper-parameter combination. After choosing the ranges of values, the NNI runs PYGON many times, selecting the hyper-parameter combinations in each experiment using a tuner (i.e. a learning tool that explores the hyper-parameter space, to find the hyper-parameter combination that gives the best performance on the model, as reported by the user). Eventually, the process converges to a neighborhood of hyper-parameters, using which PYGON performs best. Obviously the data used for the NNI were not used for reporting the results.

\section{Results}
\begin{figure*}[!t]
    \centering
    \includegraphics[width=15cm]{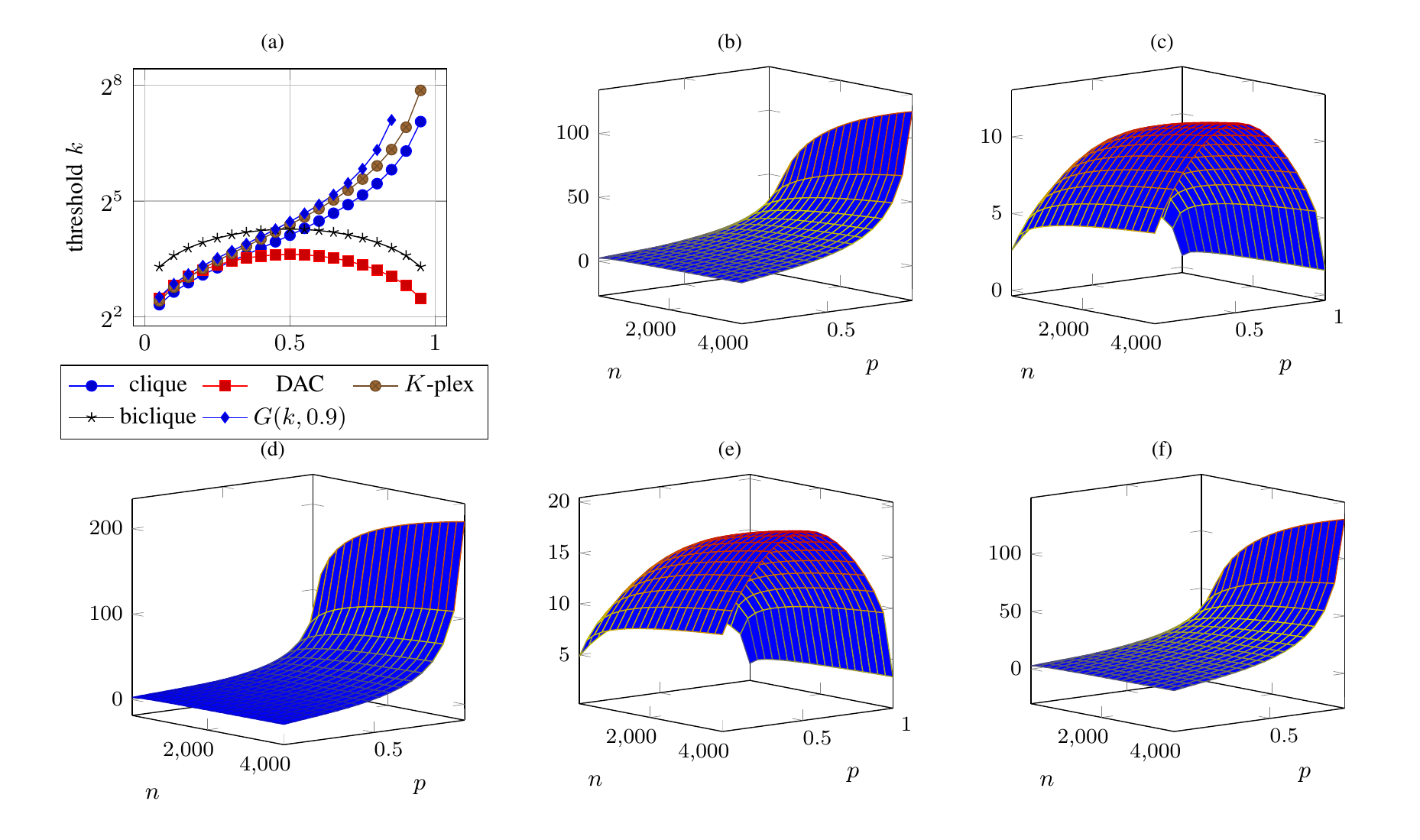}
    \caption{The theoretical minimal size $k$ of the subgraphs from \ref{Cutoff sizes}, that their expected frequency in a $G(n,p)$ would be smaller than 1, as a function of the graph size $n$ and the edge probability $p$. Fig. 2a shows all cutoff sizes for a constant graph size of 5000. The rest of the figures ((b) clique, (c) DAC, (d) $K$-plex, (e) biclique and (f) $G(k, 0.9)$) are surface plots of the cutoffs as functions $n$ and $p$. All minimal sizes are increasing as a function of the graph size. In addition, note that for Directed Acyclic Cliques and bicliques, the threshold size is not monotonic with $p$ but has a maximum for $p = 0.5$. This is because in such subgraphs, the fraction of edges between the subgraph vertices, out of the maximal possible number of edges, is around half. Hence, both having too few and having too many edges in the graph (due to small and large edge probabilities, respectively), are likely to reduce the size of a minimal subgraph of these patterns.}
    \label{fig:threshold size by prob and size}
\end{figure*}

We give an empirical evidence that PYGON can recover various dense subgraphs for sizes as small as $\Theta\left(\sqrt{n}\right)$. We study not only planted cliques, but also several other dense subgraphs, in both directed and undirected graphs. The formalism proposed for all dense subgraphs is the same, and we propose that PYGON formalism is appropriate for the recovery of any dense subgraph in a $G(n,p)$.
We study the following subgraphs:
\begin{itemize}
    \item Undirected cliques - We choose $k$ vertices and draw an edge between every pair of which (if there was no edge connecting this pair of vertices).
    \item Directed Acyclic Cliques (DAC) - Let $S$ be a set of $k$ vertices in a directed $G(n, p)$ graph. Choose a random order of the vertices: $(s_{i_1}, s_{i_2}, ..., s_{i_k})$. We call the induced subgraph $G[S]$, that its edges are all the possible edges from a vertex $s_{i_r}$ to a vertex $s_{i_t}$ where $r < t$, a \textit{Directed Acyclic Clique (DAC)}. This subgraph is a Directed Acyclic Graph (DAG) and its undirected form it is a clique.
    \item $K$-plexes - A $K$-plex is an undirected graph $H$ of size $m$ such that every vertex of $H$ is connected to at least $m-K$ vertices. Hence, it is a generalization of a $k$-clique, which can also be interpreted as a 1-plex.
    We study a specific case of $K$-plexes as our planted subgraph: a 2-plex of size $k$, such that each vertex is connected to $k-2$ vertices, maybe except for one vertex which is connected to all other vertices (to solve parity problems). 
    \item Bicliques - We plant a complete induced bipartite graph of size $k$, with sides of sizes $\lceil\frac{k}{2}\rceil$ and $\lfloor\frac{k}{2}\rfloor$. Note that this is not the known problem of recovering a hidden biclique in a random bipartite graph, because our background graph is still an undirected $G(n, p)$. 
    \item Random dense subgraph - Let $q > p$. We choose a subset of vertices $S$. We build the entire (undirected) graph such that the probability of edges between pairs of vertices from $S$ will be $q$, whereas for pairs of vertices that at least one of them is not in $S$, the edge probability will be $p$. The existence of each edge is independent on all the other edges. 
\end{itemize}
For each such graph, we compute the theoretical minimal size $k$ of the planted subgraph, that its expected frequency in a $G(n,p)$ would be smaller than 1. These sizes serve as a baseline when comparing our performance with the performance of the existing algorithms. Figure \ref{fig:threshold size by prob and size} shows the threshold sizes for the different subgraphs as functions of the graph size and edge probability.

\subsection{Cutoff for Detection of Planted Dense Subgraphs}\label{Cutoff sizes}
In the following calculations, we denote by $X_k$ the random variable enumerating the corresponding subgraphs of size $k$ in a $G \sim G(n, p)$. From the linearity of the expectation  and the independence of edges, $\mathbb{E}\left[X_k\right]$ can be calculated as:
\begin{equation}
    \mathbb{E}\left[X_k\right] = \sum_{S\subseteq V: |S| = k} \mathbb{E}\left[\mathbb{I}_{\{G[S] \cong H\}} \right] = \binom{n}{k} \mathbb{P}\left(K \cong H \right)
\end{equation}
where $\mathbb{I}_{\{R\}}$ is the indicator random variable for the occurrence of the event $R$, the subgraph is denoted by $H$, $\cong$ denotes graph isomorphism, $K$ is a random variable denoting an induced subgraph of size $k$ and the probability $\mathbb{P}$ is evaluated with respect to $G(n, p)$. 

The calculations for each subgraph case will rely only on the first moment method, hence insufficient as proof for an exact information-theoretic lower bound. However, it is sufficient for us to provide an intuition on the threshold values. Moreover, one can show that the true cutoff values are of the same order of magnitude as the ones we will present. This can be shown by letting $k$ be the value we will find in each case, times $\omega(1)$. One can easily receive that in these cases, $\mathbb{E}\left[X_k\right] = o(1)$, so using Markov's inequality, the results will be: $\mathbb{P}\left(X_k \geq 1\right) \leq \mathbb{E}\left[X_k\right] = o(1)$.

\subsubsection{Undirected Clique} \label{Clique Theory}
Matula \citeyear{Matula76} was the first to show that the threshold function for the size of a maximal clique in an undirected $G(n, p)$ graph is
\begin{equation}
    k = 2 \log_{\frac{1}{p}}{n} - 2 \log_{\frac{1}{p}}{\log_{\frac{1}{p}}{n}} + \Theta(1)
\end{equation} 
This is a stronger claim than finding the minimal $k$ for which $\mathbb{E}\left[X_k\right] \leq 1$.

\subsubsection{Directed Acyclic Cliques}
Here and in the two next cases, we only find the minimal $k$ for which $\mathbb{E}\left[X_k\right] \leq 1$. Note that the building process of the DAC indicates the number of possible options to form one in a specific set of $k$ vertices, and therefore:
\begin{equation}\label{eq:DAC threshold 1}
    \mathbb{E}\left[X_k\right] = \binom{n}{k} \cdot k! \cdot p^{\binom{k}{2}} \left(1-p\right)^{\binom{k}{2}} \approx n^k \left(p(1-p)\right)^{\frac{k(k-1)}{2}}
\end{equation}

From (\ref{eq:DAC threshold 1}), one can see that the threshold value of $k$ is:
\begin{equation}
    k = \log_{\frac{1}{\sqrt{p(1-p)}}}{n} + \Theta\left(1\right) = 2 \log_{\frac{1}{p(1-p)}}{n} + \Theta\left(1\right)
\end{equation}
\subsubsection{$K$-plexes}
We denote the size of the  $K$-plex by $m$, we randomly split the $m$ vertices we chose in advance into pairs (possibly with one vertex left alone). The number of such splits is $\frac{m !}{2^{\lfloor \frac{m}{2}\rfloor} \lfloor \frac{m}{2}\rfloor !}$, hence
\setlength{\arraycolsep}{0.0em}
\begin{equation}
\mathbb{E}\left[X_m\right] = \binom{n}{m}\frac{m!}{2^{\lfloor\frac{m}{2}\rfloor} \lfloor \frac{m}{2}\rfloor !} p^{\binom{m}{2}-\lfloor \frac{m}{2}\rfloor} (1-p)^{\lfloor \frac{m}{2}\rfloor}
\end{equation}
\setlength{\arraycolsep}{5pt}

Now, using Stirling's approximation, $2\cdot\lfloor \frac{m}{2}\rfloor\approx m$ and $\frac{m}{2} \approx \lfloor \frac{m}{2}\rfloor$:
\begin{equation}
    \mathbb{E}\left[X_m\right]\approx \frac{1}{\sqrt{\pi m}} \left[\frac{1-p}{p^2}e\cdot \frac{n^2 p^m}{m}\right]^{\lfloor \frac{m}{2}\rfloor}
\end{equation}
As a result, the threshold value of $m$ is 
\begin{equation}
    m = 2 \log_{\frac{1}{p}}{n}-\log_{\frac{1}{p}}{\log_{\frac{1}{p}}{n}} + \Theta\left(1\right)
\end{equation}
\subsubsection{Bicliques}
For calculating $\mathbb{E}\left[X_k\right]$, we will choose the $k$ vertices in two steps: first, from all the $n$ vertices choose $\lceil \frac{k}{2} \rceil$, and then, from the remaining $n - \lceil \frac{k}{2} \rceil$ choose the other $\lfloor \frac{k}{2} \rfloor$ for the second side. Therefore, up to a constant,
\begin{equation}
    \mathbb{E}\left[X_k\right] \sim \binom{n}{\lceil \frac{k}{2} \rceil} \binom{n - \lceil \frac{k}{2} \rceil}{\lfloor \frac{k}{2} \rfloor} p^{\lceil \frac{k}{2} \rceil \lfloor \frac{k}{2} \rfloor} \left(1-p\right)^{\binom{k}{2} - \lceil \frac{k}{2} \rceil \lfloor \frac{k}{2} \rfloor} 
\end{equation}
Approximating as before:

\setlength{\arraycolsep}{0.0em}
\begin{equation}
\mathbb{E}\left[X_k\right] \approx \frac{1}{2 \pi k} \left[\frac{2 e n \left(p(1-p)\right)^{\frac{k}{4}}}{\sqrt{1-p} k}\right]^k 
\end{equation}
\setlength{\arraycolsep}{5pt}

And finally, the threshold value of $k$ will be 
\begin{subequations}
\begin{align}
    k &= 4 \log_{\frac{1}{p(1-p)}}{n} - 4 \log_{\frac{1}{p(1-p)}}{\log_{\frac{1}{p(1-p)}}{n}} + \Theta\left(1\right)\\
    &= 2 \log_{\frac{1}{\sqrt{p(1-p)}}}{n} - 2 \log_{\frac{1}{\sqrt{p(1-p)}}}{\log_{\frac{1}{\sqrt{p(1-p)}}}{n}} + \Theta\left(1\right)
\end{align}
\end{subequations}
\subsubsection{Dense Random Subgraphs}
Here we plant a subgraph isomorphic to a random $G(k, q)$ in a $G(n, p)$. In this case, we do not limit ourselves to a specific subgraph pattern, hence the detection threshold for the presence of such a subgraph is not a well-defined question. Instead, we use a slightly different question and take the answer to be our baseline:
We will denote $X_k$ here as the number of subgraphs of size $k$ with at least $\binom{k}{2} q$ edges (assuming $q > p$), i.e. a density at least as large as the density a typical $G(k, q)$ is expected to have. The threshold value will be a lower bound for the intuitive cutoff value of finding "a copy of $G(k, q)$" in our $G(n, p)$ graph. 

The calculation here considers the number of choices of $k$ vertices and the number of choices of $\lceil\binom{k}{2} q\rceil$ edges that should exist between the chosen vertices. We do not care about the other edges in the subgraph. Therefore, the calculation is as follows, and we will use the same approximations we used in the other cases:
\setlength{\arraycolsep}{0.0em}
\begin{eqnarray}
\mathbb{E}\left[X_k\right]&{}={}&\binom{n}{k} \binom{\binom{k}{2}}{\lceil\binom{k}{2} q\rceil} p^{\lceil\binom{k}{2} q\rceil} \nonumber \\ &&{}\:
    \approx 2\pi k\sqrt{q\frac{k-1}{2}} \left[\frac{n e}{k} \left(\frac{e p}{q}\right)^{\frac{k-1}{2} q} \right]^{k}
\end{eqnarray}
\setlength{\arraycolsep}{5pt}
Again, the threshold order of magnitude of $k$ is logarithmic with $n$:
\begin{equation}
    k = \frac{2}{q}\log_{\frac{q}{p}}{n} - \frac{2}{q}\log_{\frac{q}{p}}{\log_{\frac{q}{p}}{n}} + \Theta\left(1\right)
\end{equation}

\subsection{Modifications in the Adjacency Matrix}
\label{3.2}
As mentioned in section 3.2.3, our choice of adjacency matrix was not trivial. In this subsection, we describe how our choice affects the performance of PYGON. 
\subsubsection{Adjacency Matrix with Learnable Parameters}
The choice of our adjacency matrix had been made after comparing several common variations of adjacency matrices:
\begin{itemize}
    \item $\tilde{A}=D^{-\frac{1}{2}} (A+A^T+I) D^{-\frac{1}{2}}$, where $D$ is a diagonal matrix of the vertices' degrees. This is a common use of adjacency matrix in learning tasks (see \cite{KW16} for example), however in our case this choice has some drawbacks. The most important of them is the lack of effect of an absence of edges between neighbors. We found that in our type of problems, missing edges are crucial for learning.
    \item $\tilde{A}=D^{-\frac{1}{2}} (W+W^T+I) D^{-\frac{1}{2}}$, where $W = 2A - \mathds{1}$ and $\mathds{1}$ is a matrix whose elements are all 1. Using this matrix, the model learned fine, however there was a room for improvement. First, we wanted the model to control the effect of self loops, so we put a learnable coefficient $\gamma$ multiplying $I$ in the matrix, which had small but positive effect. 
    \item Then, we looked for more expressiveness in the messages passing through the edges. Therefore, we set the weights of edges to be some learnable coefficient $\alpha$, and the weights of missing edges to be another learnable coefficient $\beta$. In order to keep the model from having too many degrees of freedom and so to not learn, we kept keep $\alpha$ constant. However, this was not enough, since we saw in training that $\alpha$ and $\beta$ could eventually get similar signs. This way, the performance was damaged because having or not having an edge between two vertices was not different enough.
    \item Eventually, to keep the weights of edges and missing edges with different signs, we set the edge weights to be $e^\alpha$ and the missing edges to be $-e^\beta$, as in (\ref{eq:adjacency matrix}).
\end{itemize} 

The choice of learnable parameters gave freedom to learn the global effect of having an edge comparing to lacking one, while the specific form of exponents limits the possibilities of edge weights to the correct sign, allowing us to express some prior knowledge regarding the effect of the existence or absence of edges. 

\subsubsection{In Dense Graphs, GCN Must Include Penalty for the Absence of Edges}

\begin{figure}[!t]
    \centering
    \includegraphics[width=10.5cm, height=9cm]{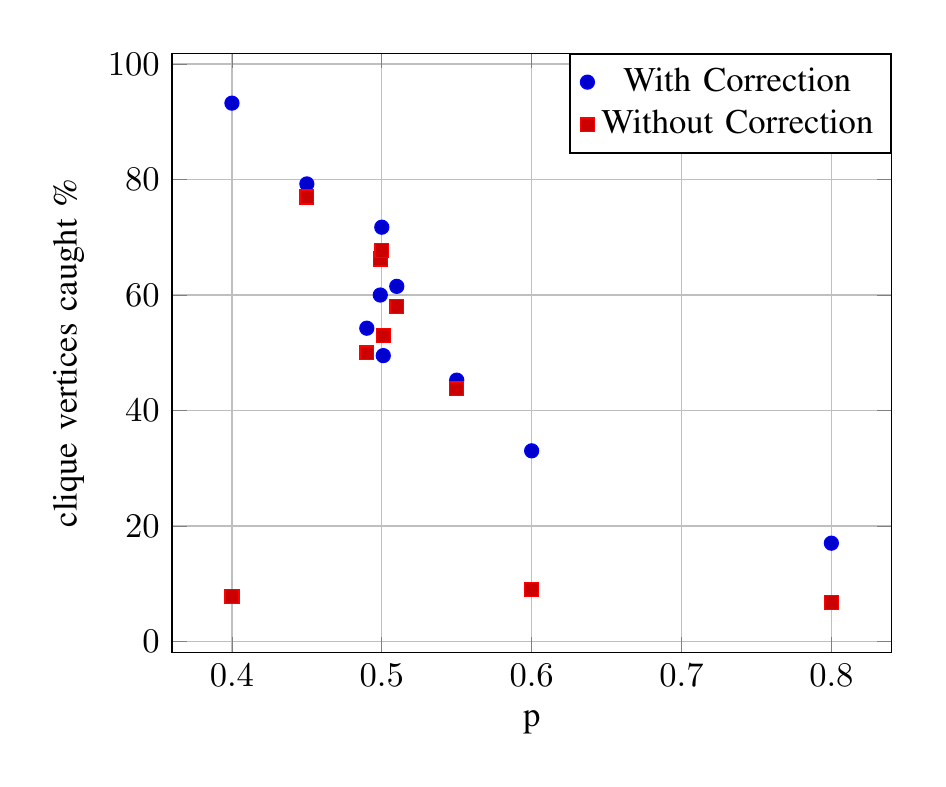}
    \caption{The average percentages of clique vertices caught among the top 40 vertices by PYGON scores, in $G(500, p, 20)$ graphs, for different values of $p$, with and without correcting the edge weights. We show here more values of the edge probability as the probability approaches 0.5, i.e. the value where the edge correction has no effect.}
    \label{fig:edge correction}
\end{figure}

GCN are often used for the  embedding of the graph vertices. To create this embedding, the GCN layer uses both the existing feature representation and the graph topology, input to this layer as $H^{(l)} \in \mathbb{R}^{n\times f_l}$ and $\Tilde{A} \in \mathbb{R}^{n\times n}$ respectively, such that each vertex will be represented as:
\begin{equation}
    H_{i, 1:f_{l+1}}^{(l+1)} = \sigma\left[\sum_{m}{\left(\sum_{k}{\tilde{A}_{i,k} H_{k,m}^{(l)}}\right) W^{(l+1)}_{m, 1:f_{l+1}}}\right]
\end{equation}
Note that the term in the inner parentheses determines what vertices are used in order to build the new representation of each vertex, and how. For example, if $\tilde{A}$ is simply $A$ (or $A+A^T$ for directed graphs), then the vertex representation is built from the previous representations of its first neighborhood, and if $\tilde{A} = A+I$ (or $A+A^T + I$ for directed graphs) then the neighbors and the vertex itself are used for the new representation. 

GCN are typically used in real-world graphs, which are much sparser than  $G(n, p)$ graphs for $p$ close to 0.5. Hence, the embedding of vertices in these graphs is built based on a small number of vectors, since the neighborhoods of the vertices are small. However, in dense graphs, the embedding is built based on $\Theta\left(n\right)$ vertices. In addition, in the task of detecting dense subgraphs, the absence of edges is of great importance, therefore we use $\tilde{A}$ with negative weights in entries corresponding to missing edges (i.e. if $A_{i, j} = 0$, then $\tilde{A}_{i, j}$ is negative), and the output vector of each vertex is built based on the input vectors of all vertices.

Putting negative weights on missing edges creates a problem: considering a $G(n,p)$ instance for $p \neq \frac{1}{2}$, the GCN layer creates a vector representation that is biased due to the difference between the number of positive and negative contributions. This bias, if not treated, fails the training process. Moreover, PYGON contains several GCN layers and the bias grows exponentially with the number of layers. We solve the bias problem when using PYGON on graphs with $p\neq \frac{1}{2}$, by  multiplying the weights of the positive edges in $\tilde{A}$ by a constant $\mu$, such that for a random vertex $i\in V$:
\begin{equation}
    \mathbb{E}\left[\mu \cdot |\mathcal{N}\left(i\right)| - |V\setminus\mathcal{N}\left(i\right)|\right] = 0
\end{equation}
where $\mathcal{N}\left(i\right)$ is the neighborhood of $i$, leading to: $\mu = \frac{1-p}{p}$, as in (\ref{eq:adjacency matrix}). Setting the initial edge weights by $\mu$ sets an unbiased starting point, from which the model can be trained to adapt $\mu$ to the task in hand.

Figure \ref{fig:edge correction} shows the importance of using this correction. The expected bias: \\* $\mathbb{E}\left[\mu \cdot |\mathcal{N}\left(i\right)| - |V\setminus\mathcal{N}\left(i\right)|\right]$, grows with the deviation from $p=0.5$, and the performance of PYGON without the correction decreases dramatically, since training the model becomes practically impossible.

In every test of PYGON, we tried to find a planted subgraph of known pattern and size\footnote{The values of $k$ we used were large enough comparing to the cutoff value, so that the planted subgraphs were almost surely the only subgraphs of that patterns and sizes.}, in $G(n, p)$ graphs with specific, known $n, p$. We initialized 20 random realizations (i.e. 20 $G(n, p)$ graphs with planted subgraphs of the desired size) per instance, separated them into 5 equal folds and ran 5 cross-validations, such that every run, 3 folds (i.e. 12 graphs) were used as training graphs, 1 as validation and 1 as test. The performance was measured by averaging the desired measurement on the 20 graphs when they were used as test graphs.

\subsection{Clique Recovery}

\begin{figure}[!t]
\centering
    \includegraphics[width=10.5cm, height=9.5cm]{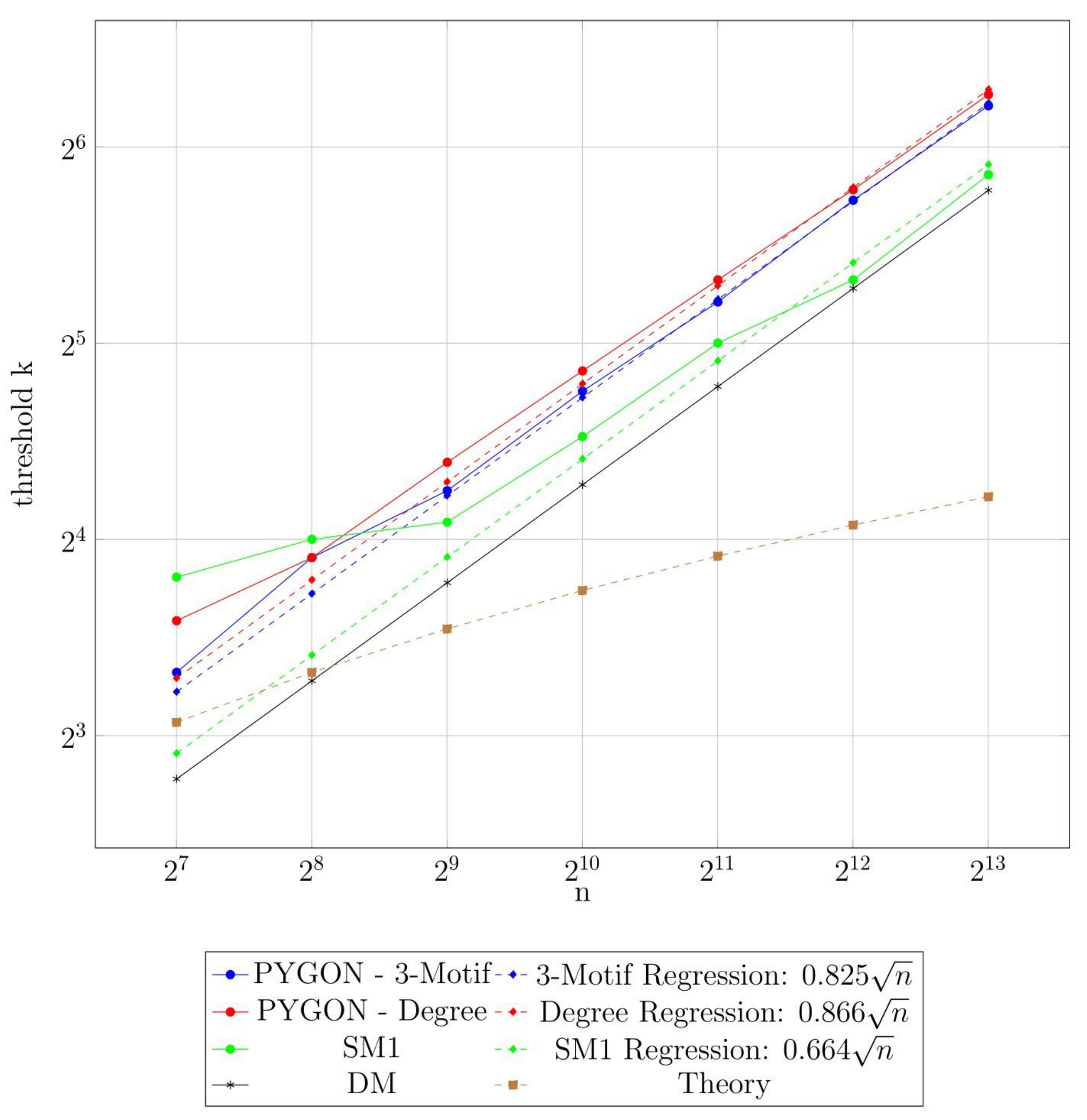}
    \caption{Threshold sizes $k$ of cliques, for which at least 50\% of the clique was found in the top $2k$ scores from PYGON, as a function of the graph size $n$. We show the performance of PYGON, either using 3-motifs or degrees as the initial features per vertex. We compare PYGON with the theoretical value calculated in \ref{Clique Theory}, the theoretical performance shown by Deshpande and Montanari \citeyear{DM15}, represented here as DM, and experimental results of $SM^1$ algorithm. Note that the theoretical value calculated in \ref{Clique Theory} applies for large values of $n$. A regression of the performance of PYGON to the form $\alpha \sqrt{n}$ is also shown, finding that using 3-motifs, $\alpha \approx 0.825$ and using the degree, $\alpha \approx 0.866$. The algorithm of DM can find cliques of size $\sqrt{n/e} \approx 0.606\sqrt{n}$ in theory, and a regression of the performance of $SM^1$ algorithm to the form $k=\alpha\sqrt{n}$ has found that $\alpha\approx 0.664$.}
    \label{fig:threshold cliques}
\end{figure}

PYGON outputs a test score for each vertex in the test graphs. Using this score, one can order the vertices from each graph, where the vertices that form the planted clique should be the top vertices (the vertices with the highest scores, from this model). Choosing the top score vertices is the step before the cleaning/extending algorithm, that should give the final set of vertices.

Note also that the framework of PYGON can be used with a graph neural network layer other than GCN, as demonstrated with Graph Attention Network (GAT) layer \cite{VCC+17} in appendix C. The framework can be used with other, more general neural network layers, such as feedforward or convolutional neural networks, however the models we tried using those layers failed to learn. 

We compared the results of PYGON with the results claimed by Deshpande and Montanari \citeyear{DM15}, later addressed as DM, and the results of Marino and Kirkptrick's $SM^1$ algorithm \citeyear{MK18}. Note that the algorithm of DM is of time complexity of $O(n^2 \log{n})$ and $SM^1$ is of time complexity of $O(n^3)$, comparing to PYGON, which can have time complexity as short as $O(n^2)$ (more about PYGON's time complexity can be found in \ref{time complexity}). 
To compare the performance of PYGON to DM and $SM^1$ in clique recovery, we generated instances of graphs of sizes 128 to 8192 with equal spaces in logarithmic scale (base 2). We looked for the value $k$ for which PYGON recovers 50\% of the planted clique, out of the $2k$ vertices with the highest scores. The hyper-parameters for our model are specified in appendix A. The results are shown in Figure \ref{fig:threshold cliques}. A simple regression shows that PYGON can recover cliques of sizes as low as $0.825\sqrt{n}$ using 3-motifs as the initial features, and as low as $0.866\sqrt{n}$ using the degree as the initial feature. The reason why we chose $2k$ and top 50\% is that using a cleaning procedure similar to DM, we have reached success rates of roughly 50\% as well.

\subsection{Recovery of Other Dense Subgraphs} 
\begin{figure*}[!t]
    \centering
    \includegraphics[width=15cm, height=12cm]{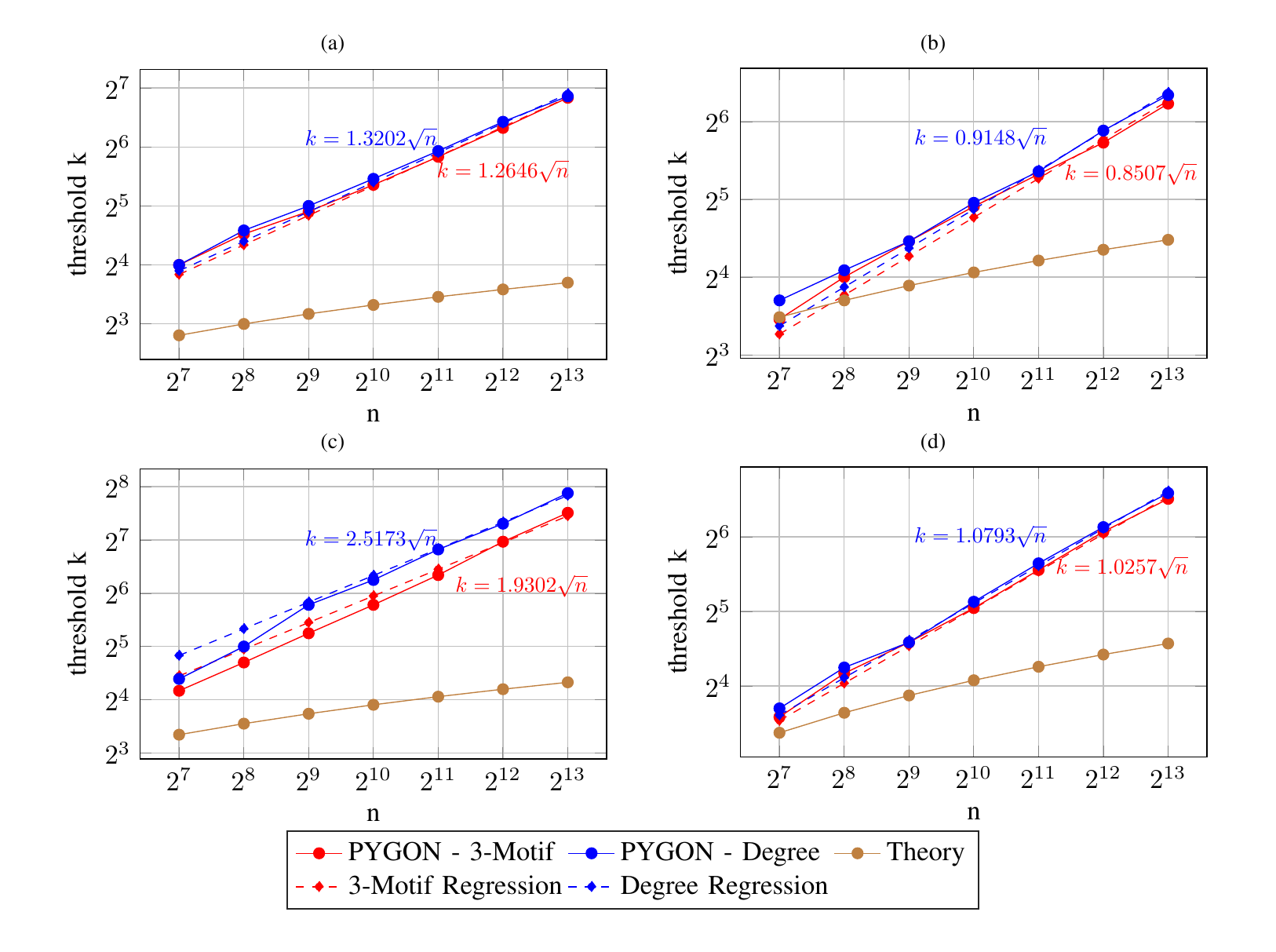}
    \caption{Threshold sizes $k$ of the studied subgraphs ((a) DAC, (b) $K$-plex, (c) biclique and (d) $G(k, 0.9)$), for which at least 50\% of the subgraph was found in the top $2k$ scores from PYGON (either using 3-motifs or degrees as the initial features), as a function of the graph size $n$ (we used $n=128, 256, ..., 8192$). Here we compare the performance of PYGON with the theoretical value calculated in \ref{Clique Theory}. Note that for bicliques, we use $p = 0.4$, whereas for the other subgraphs we use $p = 0.5$. In each plot we added the regression from PYGON values to the form $\alpha \sqrt{n}$ and the corresponding equations.}
    \label{fig:threshold subgraphs}
\end{figure*}

As mentioned previously, PYGON is capable of recovering various dense subgraphs without adaptations for specific subgraphs. Using the exact same framework (only trained on the relevant subgraph each time) and hyper-parameters, we tested the threshold value of $k$ for which we find at least half the subgraph in the top $2k$ vertices by model scores, for the subgraphs presented above. The results are shown in Figure \ref{fig:threshold subgraphs}, where we present PYGON performance with either 3-motifs or degrees as the initial features. Indeed, PYGON can recover the subgraphs presented above, for sizes $k = \Theta\left(\sqrt{n}\right)$, namely the performance of PYGON is not affected by the type of the subgraph required to recover.

\begin{figure*}[!t]
    \centering
    \includegraphics[width=15cm]{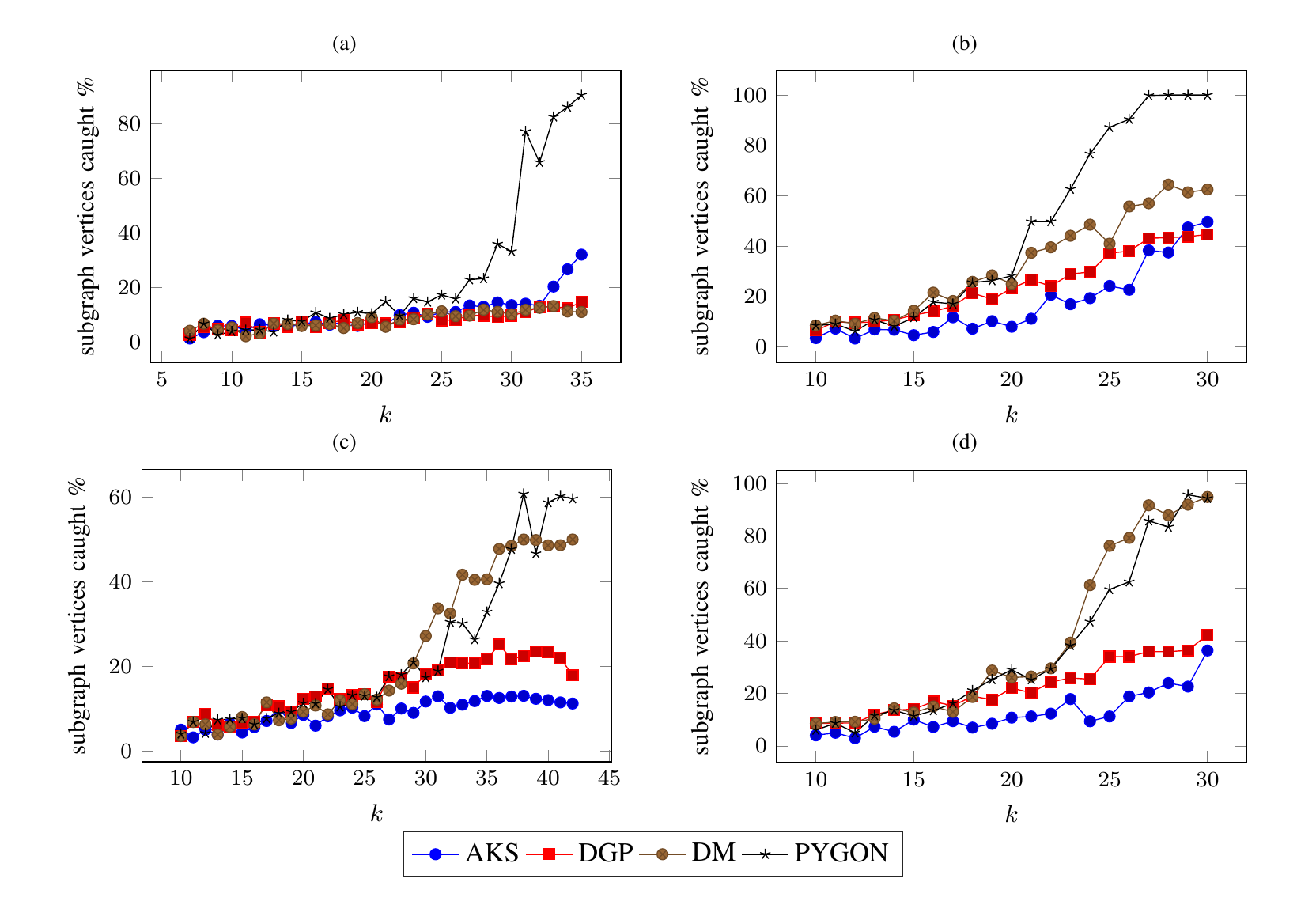}
    \caption{Average percentages of subgraph vertices captured in the top $2k$ vertices as a function of $k$, for graphs of size 500 with planted subgraphs ((a) DAC, (b) $K$-plex, (c) biclique and (d) $G(k, 0.9)$) of varying sizes. The edge probabilities are 0.5 for all graphs, except for the biclique, where the probability is 0.4. We compare the performance of PYGON with the performances of existing algorithms \cite[for AKS, DGP and DM, respectively]{AKS98,DGP14,DM15}.}
    \label{fig:comparison}
\end{figure*}

As no other known algorithm can perform on both cliques and other subgraphs, we also compare the performance of PYGON to other clique recovery algorithms \cite{AKS98,DGP14,DM15}. In Figure \ref{fig:comparison}, one can see that the only model which is not affected by structure of the planted subgraph and by the question whether the graph is directed is PYGON.

\subsection{Time Complexity and Memory Cost}\label{time complexity}
The theoretical time complexity of our algorithm is affected mainly by the feature calculations and GCN layers. As shown in \ref{Features per Vertex}, the time complexity of calculating 3-motifs as initial features is $O(n^3)$, whereas using degrees as initial features, the complexity decreases to $O(n^2)$. According to Kipf and Welling \citeyear{KW16}, the time complexity of a GCN layer is $O(n^2\cdot f_{i}\cdot f_{i+1})$ where $f_{i}$, $f_{i+1}$ are the input and output feature dimensions, respectively. Therefore, the total time complexity of PYGON with 3-motifs as initial features is $O(n^3)$, and with degrees as initial features, the complexity is $O(n^2)$.

Recall that the time complexity of PYGON with degrees as initial features is of the same order of magnitude as the algorithm of Dekel et al. \citeyear{DGP14}, and is lower than the algorithm of DM \citeyear{DM15}. Even in case we use 3-motifs as initial features, the use of GPU in our algorithm (in both feature calculations and in neural networks) shortens the computation time substantially, such that the running times of PYGON and some existing methods are similar (42 sec for DM vs less than 50 for PYGON, for a test of 20 graphs).
Note that PYGON still requires more time to train, a stage that tends to be longer than the other algorithms, however once the model is trained, any new graph can be fed into the trained model without additional actions.

\begin{figure}[!t]
    \centering
    \includegraphics[width=8cm]{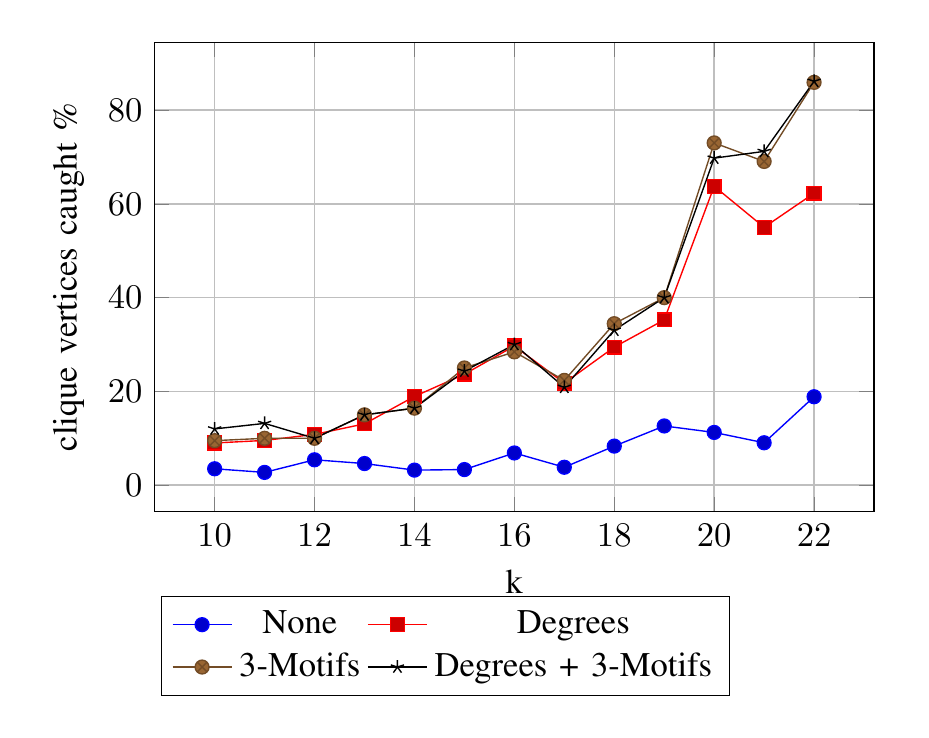}
    \caption{Average percentage of clique vertices captured in the top $2k$ vertices as a function of $k$, for $G(500, 0.5)$ graph with planted clique of varying sizes, using different sets of initial features. We show here that using initial features of lower complexity may have only a minor effect on the performance of PYGON comparing to higher complexity features. In addition, the use of PYGON without initial features is possible, although not as effective as low-complexity features.}
    \label{fig:initial features}
\end{figure}

As mentioned in \ref{Features per Vertex}, one can choose using different initial features, including the option not to calculate features at all (in this case, each vertex is represented initially by a one-hot vector). The option of using cheap features should be considered, since the feature calculations take the majority of the computation time PYGON needs (especially when using features costing $O(n^3)$ in time). Figure \ref{fig:initial features} shows the performance of PYGON using different sets of initial features. One can see that the initial features affect the performance, although not by much, hence saving time using PYGON with less expensive initial features (e.g. degrees only, or scores from few iterations of belief propagation) is possible. 

The main memory limit of PYGON is the cost of storing the adjacency matrix. Since the graphs studied here are $G(n,p)$ graphs, the memory required to hold the adjacency matrix is $O(n^2)$. Typical GPUs have a memory limit of around 1-10GB. Hence, running PYGON is limited to graphs of 10,000-20,000 vertices (i.e., 100,000,000-400,000,000 edges), depending on the GPU limitations. 

\subsection{Extension of Subgraph Candidates to Full Subgraph} 
In many existing algorithms, the last one or few stages are intended to take the remaining set, in which are many vertices that belong to the planted subgraph, clean and extend it to recover the whole subgraph. Using the outputs of PYGON, one can apply a similar stage, with any desired cleaning algorithm, to find the planted subgraph. 

In our use, we took the set of candidates before cleaning to be the $2k$ vertices that received the highest scores from PYGON. The selection of $2k$ vertices has two reasons. On one hand, selecting too many vertices increases the computational cost of the cleaning algorithm. On the other hand, selecting too few vertices has a risk of losing too much signal, causing the algorithm to fail in finding the entire subgraph. We tested and found that the appropriate size of set of candidates for the algorithm we used is $2k$. 

Specifically, we used the methods described in Algorithm \ref{alg:Cleaning Algorithm}, which is similar to the cleaning algorithm presented in DM \citeyear{DM15}, on the task of recovering the clique. We tried several variations of algorithms similar to the algorithm presented in DM, then selected the best variation. Algorithm \ref{alg:Cleaning Algorithm} has two main differences from the cleaning algorithm of DM. First, when we selected candidate vertices, we took a constant number (with $k$), whereas DM selected all the candidates that meet some criteria, i.e., a number that can change. We did not select candidates by criteria because, unlike DM, we do not have a theoretic base to rely on when having our candidates, hence could not select criteria that guarantee us enough vertices. Second, we applied the second part of the algorithm, where we select vertices by connectivity, in a loop instead of one time only. We saw that this loop improves the performance comparing to running only once. 

\begin{figure}[t]
    \centering
    \includegraphics[width=15cm]{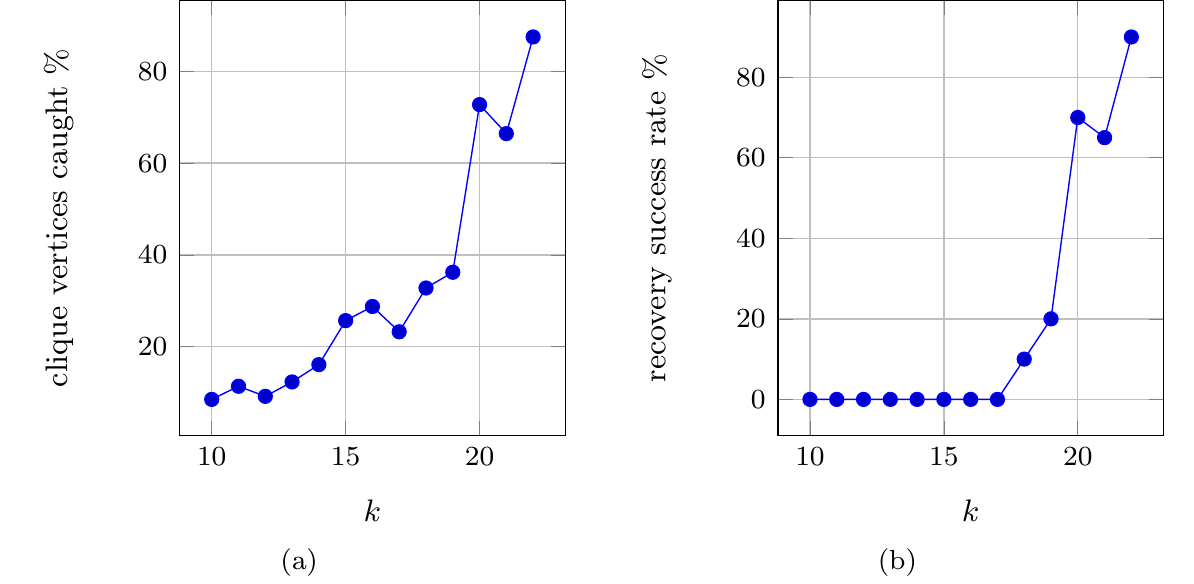}
    \caption{The performance of PYGON on planted cliques in $G(500, 0.5)$ graphs. Figure 8(a) shows the average percentage of clique vertices caught in the top $2k$ vertices by PYGON. 
    Figure 8(b) shows the average success rates of Algorithm \ref{alg:Cleaning Algorithm} to recover the planted clique from the candidates found earlier.}
    \label{fig:cleaning procedure}
\end{figure}

This algorithm is applied graph-wise, independent on all other graphs. Figure \ref{fig:cleaning procedure} shows the recovery rate of the algorithm used on the outputs of PYGON, comparing to the percentages of vertices that belong to the clique and caught by PYGON. One can see that the performances are very similar, and that indeed when PYGON succeeds to find at least half the clique among the $2k$ proposed candidates, the cleaning algorithm shows almost the exact same performance on the second task of fully recovering the clique.

\begin{algorithm}
\caption{Cleaning Algorithm}
\label{alg:Cleaning Algorithm}
\begin{algorithmic}[1]
\STATE{\textbf{Input:} A set $S$ of candidates, clique size $k$.}
\STATE{$\Bar{A} = A[S]$, the adjacency matrix of the subgraph induced by $S$.}
\STATE{$\Hat{A} = \Bar{A} + \Bar{A}^T - \mathds{1} + I$, where $\mathds{1}$ is a matrix whose elements are all 1.}
\STATE{Compute the eigenvector $u$ corresponding to the largest eigenvalue of $\Hat{A}$.}
\STATE{Compute the set of $k$ vertices with the largest entries in $v$ by absolute value, $T^*$.}
\WHILE{$T^*$ is not a $k$-clique, or enough iterations have been done}
\STATE{For every $v\in V$, Compute $n_v = |\{w\in T^*: v,w\text{ are (weakly) connected}\}|$.}
\STATE{The new $T^*$ will be the set of the $k$ vertices with the largest sets $n_v$.}
\ENDWHILE
\RETURN $T^*$, a clique of size $k$, or failure.
\end{algorithmic}
\end{algorithm}

\section{Conclusions}

We presented PYGON -- a novel method to recover planted dense subgraphs in $G(n,p)$ graphs. The framework of PYGON makes use of known elements from graph based machine learning and graph algorithms all combined to a model for recovering dense subgraphs. The novelty of this approach is the point of view on the problem of recovering subgraphs -- the model is trained on several realizations of the problem\footnote{The training on similar realizations of the problem is crucial. When training PYGON using realizations of one problem and testing on a different problem, the performance will decrease significantly, comparing to training and testing on the same problem.} with known planted subgraphs, creating a tool that can help recovering planted subgraphs of any type and in any type of network. 

The learning process of PYGON involves the commonly used GCN layers, which pass messages through the graph to embed each vertex by the topology of the graph. The idea of passing messages between vertices had already been used by Deshpande and Montanari \citeyear{DM15}, with two main differences between their algorithm and PYGON: First, PYGON uses a limited fixed number of message passes (i.e. a finite number of GCN layers), and second, the messages are not constant from one pass to another (not even in the embedding dimension between iterations). We have shown that, despite passing messages for fewer iterations and using an algorithm of lower time complexity, our performance is close to the performance of DM \citeyear{DM15}.

Although the focus here was on the planted clique problem, PYGON works for different planted subgraphs, in both directed and undirected graphs, unlike any other known algorithm. Moreover, the existing methods are built to detect density, but not a specific structure of a planted subgraph, at least in the initial recovery of the subgraph core. In contrast, PYGON trains using realizations of the specific dense subgraph it aims to find. 

PYGON is not built to solve some other known subgraph recovery problems, such as recovering a planted subgraph in a bipartite graph (e.g. the planted biclique problem), the k-densest subgraph, the maximal common subgraph and recovering a planted non-dense specific subgraph. In addition, we have tried PYGON only in cases where a single subgraph had been planted, and did not address cases of existing hint vertices, as done by Marino and Kirkpatrick \citeyear{MK18}. We believe that PYGON can be extended to most of these problems, if not all of them, and may be able to help in the cases of multiple planted subgraphs or known partial information on the planted subgraph(s). Such extensions would require changing the structure of the learning layers (which are currently a specific implementation of a GCN layer). 

We have shown that PYGON, like all other existing approaches, does not break through the threshold of $k=\Theta\left(\sqrt{n}\right)$, for any of the dense subgraphs. A simple explanation for that would be that the classification of each vertex is the result of a combination of topological input variables. The standard deviation of the degree is proportional to $\sqrt{n}$. One can assume that all other topological features have a variance as least as large. To resolve the ambiguity between vertices belonging to the planted dense subgraph and those not belonging to it, one would need the subgraph to be as large as the expected standard deviation of the measure in non-subgraph vertices. This may suggest the hypothesis that recovering planted subgraphs of smaller sizes might be impossible in polynomial time. We propose the following conjecture:
\begin{conjecture}
    Let $G$ be a $G(n, p)$ graph, for $p$ independent on $n$, where we plant a subgraph of size $k$. Then, if $k$ is of size smaller than $O\left(\sqrt{n}\right)$, then there is no PAC-learning algorithm in a fixed degree polynomial time, that can detect the planted subgraph.
\end{conjecture}

Lastly, we used here 3-motifs as input features. One may think of this usage as a step further from Ku\v{c}era's \citeyear{Kucera95} use of degrees (i.e. 2-motifs) to detect cliques of size $k = \Omega\left(\sqrt{n \log{n}}\right)$. We believe that for planted cliques and possibly other subgraphs, of any known size $k$ larger than the detection threshold of the subgraph, one can find an appropriate number $s$ (significantly smaller than $\Theta\left(\log_{\frac{1}{p}}{n}\right)$, but probably dependent on $n$), such that the use of $s$-motifs can help recovering the planted subgraph with high probability and in quasi-polynomial time, shorter than the known exhaustive search algorithm. For now, this question is left unanswered.

\appendix
\section*{Appendix A. Model Hyper-Parameters}\label{Appendix A} 
The following PYGON hyper-parameters were found using Microsoft Corporation's NNI \citeyear{NNI}:

In most figures, we take the initial feature matrix to be built from 3-motifs (hence, for undirected graphs, it has 2 columns, and for directed graphs, it has 6 columns). Several specific figures also include PYGON models with only degrees as the input features. Either input matrix is normalized by taking a logarithm (more precisely, $\log_{10}{\max{\left(10^{-10}, x\right)}}$) and then z-scoring over the columns (on all the graphs together). The learned model is composed of 5 GCN layers, with hidden layers of sizes 225, 175, 400 and 150 (the input dimension is the number of features and the output dimension is 1). The dropout rate is 0.4. The model is trained for 1000 epochs at most (due to early stopping, the training process stops much earlier, after 250 epochs at most). The optimizer is ADAM (with the default hyper-parameters of the Pytorch implementation), with learning rate of 0.005 and $L_2$ regularization coefficient of 0.0005.

\appendix
\section*{Appendix B. Best Tested Loss Function Definition}\label{Appendix B}
In order to try and find a preferable loss function, we used a more general definition of loss function as the function to minimize in the training process of PYGON:
\begin{equation}\label{eq:extended loss}
    L\left(\hat{y},\ y\right) = WBCE\left(\hat{y},\ y\right) + c_1 P\left(A, \hat{y}\right) + c_2 B\left(\hat{y}\right)
\end{equation}

Where:
\begin{itemize}
    \item $A$ is the adjacency matrix ($A_{i,j} = 1$ if $\{i, j\}\in E$ and 0 otherwise).
    \item $ P\left(A, \hat{y}\right) = - \frac{1}{n^2}\sum_{i,\ j\in V}$\\ $\left[A_{i,j}\log{\left(\frac{1+{\hat{y}}_i{\hat{y}}_j}{2}\right)}+\left(1-A_{i,j}\right)\log{\left(\frac{1-{\hat{y}}_i{\hat{y}}_j}{2}\right)}\right]$\\ is a pairwise loss for the clique recovery, as used by Chiara \citeyear{Angelini18}. 
    \item $B\left(\hat{y}\right) = -\frac{1}{n} \sum_{i\in V}\left[{\hat{y}}_i\log{\frac{k}{n}} +\left(1-{\hat{y}}_i\right)\log{\left(1-\frac{k}{n}\right)}\right]$ is a binomial regularization, as used by Chiara \citeyear{Angelini18}. 
    \item $c_{1}, c_{2}$ are constants. 
\end{itemize}

The first term is a weighted binary cross-entropy function. We used weights to correct the severe imbalance between the positive and negative samples (i.e. vertices from the planted subgraph and vertices that do not belong to the planted subgraph, respectively).

The two last terms are additions to the loss function as used by Chiara \citeyear{Angelini18}, specifically for the task of recovering cliques. The pairwise loss forces that two model scores for non-adjacent vertices will never be both 1, and that two adjacent vertices will both have high scores (though the penalty of the latter is not as high as the former). The binomial regularization keeps the number of vertices predicted to be clique vertices from being too large.

\begin{figure}[!t]
    \centering
    \includegraphics[width=10cm]{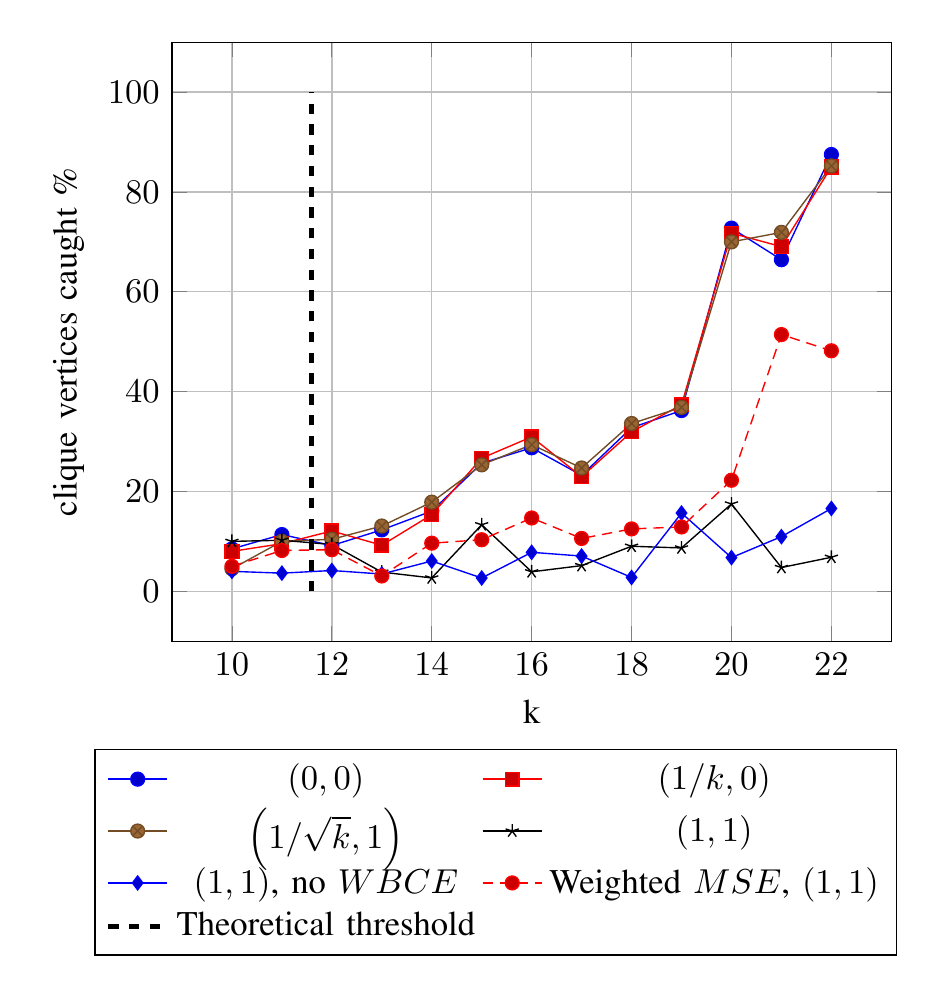}
    \caption{Percentages of vertices from the planted clique in the top $2k$ scores given by PYGON, as a function of the planted clique size $k$. We used different loss functions, depending on the coefficients $\left(c_{1}, c_{2}\right)$ in (\ref{eq:extended loss}), aiming to recover planted cliques in $G(500, 0.5)$ graphs. We measure the performance with and without adding the pairwise loss and binomial regularization to the $WBCE$ loss. We also test the effect of the $WBCE$ loss comparing to weighted $MSE$ loss (same sample weights as in $WBCE$) instead and to no loss instead of $WBCE$. The theoretical size of the largest cliques in $G(500, 0.5)$ graphs without planted cliques is shown as well.}
    \label{fig:different losses}
\end{figure}

In PYGON, the pairwise loss and binomial regularization are either not effective or damage the performance, whereas the weighted binary cross-entropy is crucial, as shown in Figure \ref{fig:different losses}. 

\appendix
\section*{Appendix C. PYGON with GAT Layer}\label{Appendix C}

\begin{figure}[!t]
    \centering
    \includegraphics[width=10cm]{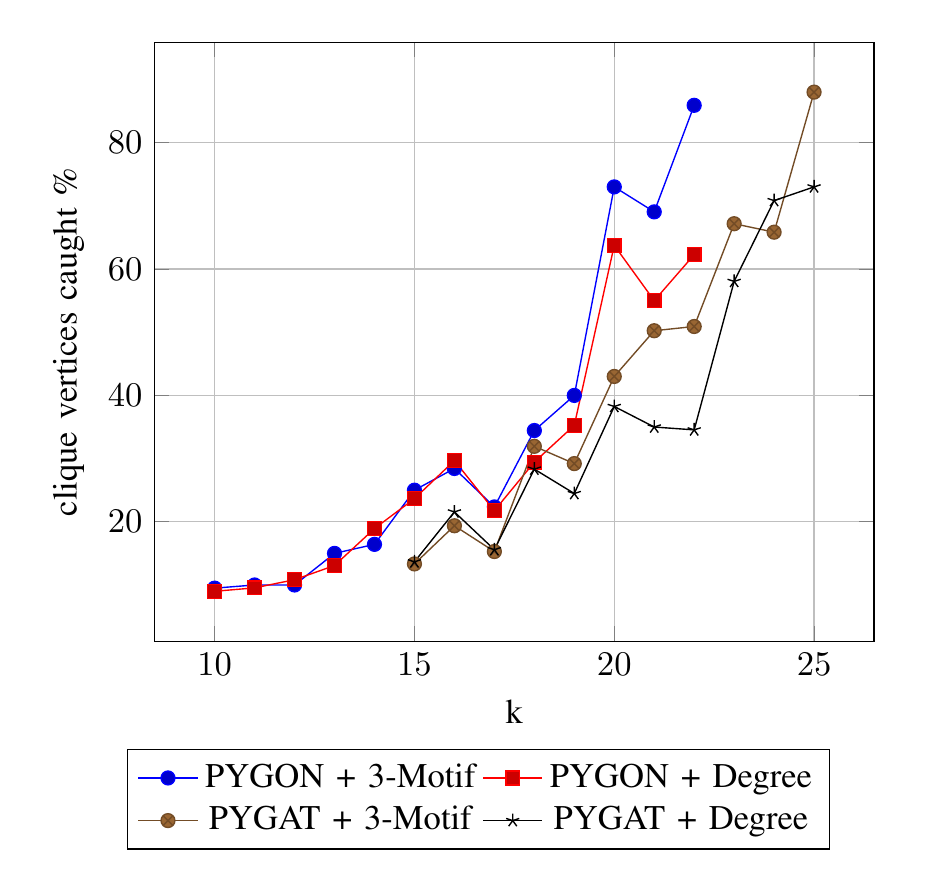}
    \caption{Average percentages of vertices in the top $2k$ predicted scores, that belong to the planted clique, as a function of the planted clique size $k$ (in $G(500, 0.5)$ graphs). We compare here the predicting models: PYGON vs. PYGAT, each with either 3-motifs or degrees as initial features. Note that for GCN-based models, we run over clique sizes 10 to 22, whereas for GAT-based models, we run over clique sizes 15 to 25}
    \label{fig:PYGON vs GAT}
\end{figure}

In order to demonstrate another level of flexibility in PYGON, we will show here that when replacing the learning layer from GCN to Graph Attention Network (GAT) layer, the performance of the algorithm (denoted as PYGAT) is similar. GAT, is a graph neural network layer that receives as input a feature representation of the vertices of a graph, and finds a new embedding of the vertices by message passing. This is done by applying a self-attentional mechanism, to learn the messages passed over each edge of the graph. 

Here, we use a specific implementation of this layer, aiming to pass information between existing neighbors, and (different) information between non-adjacent vertices: The input feature matrix is passed through a dropout layer, then a linear layer. Two learnable parameters are matched to each vertex, one for the vertex as a source vertex and one for that vertex as a target. From the inner product of the feature matrix and the learnable parameters, which is concatenated (scores of sources to the scores of targets), passed in LeakyReLU, softmax and then dropout, an attention coefficient is obtained to each edge. Then, the features are multiplied element-wise by the attention coefficients and we sum, for each vertex, over the results of its neighbors, to make the output features. Before passing to the next layer, a skip connection is added to those features. 

Our version uses two-headed attention -- one based on the original graph, and the other on the complement graph, such that the output features from this layer are a concatenation of the output features from each attention head. Only the output layer is a GAT layer based only on the original graph. 

The hyper-parameters we used in PYGAT are the following:

The initial feature matrix is built from either 3-motifs or degrees, normalized exactly like in PYGON. The learned model is made of 4 GAT layers, with hidden layers of sizes 60, 120, 170 and 100. The dropout rate is 0.05. The model is trained for 1000 epochs at most, with the same early stopping mechanism as in PYGON. The optimizer is SGD (with the default hyper-parameters of the Pytorch implementation), with learning rate of 0.13 for the 3-motif runs and 0.175 for the degree runs, and $L_2$ regularization coefficient of 0.0025 for the 3-motif runs and 0.005 for the degree runs.

Figure \ref{fig:PYGON vs GAT} shows the performance of PYGON and PYGAT. As one may see, PYGON performs slightly better, however the differences are small.

\bibliography{bibliographyjair.bib}
\bibliographystyle{theapa}

\end{document}